\documentclass[a4paper,11pt]{article}
\usepackage[a4paper, total={6.5in, 9in}, left=1in, right=1in]{geometry}

\usepackage{moreverb,url}
\usepackage[title]{appendix}
\usepackage{amssymb}
\usepackage{amsmath}
\usepackage{amsfonts}
\usepackage{bm}
\usepackage{mathtools}
\usepackage{authblk}
\usepackage{natbib}
\usepackage{wasysym}
\usepackage{svg}
\usepackage{float} 
\usepackage{graphicx}
\usepackage[colorlinks,bookmarksopen,bookmarksnumbered,citecolor=red,urlcolor=red]{hyperref}

\newcommand\BibTeX{{\rmfamily B\kern-.05em \textsc{i\kern-.025em b}\kern-.08em
T\kern-.1667em\lower.7ex\hbox{E}\kern-.125emX}}

\title{A Physically Consistent Stiffness Formulation for Contact-Rich Manipulation}

\author[1,2]{Johannes Lachner}
\author[1]{Moses C. Nah}
\author[1,2]{Neville Hogan}

\affil[1]{Department of Mechanical Engineering, Massachusetts Institute of Technology, Cambridge, USA}
\affil[2]{Department of Brain and Cognitive Sciences, Massachusetts Institute of Technology, Cambridge, USA}

\date{\today}

\begin{document}

\maketitle

\begin{abstract}
Ensuring symmetric stiffness in impedance-controlled robots is crucial for physically meaningful and stable interaction in contact-rich manipulation. Conventional approaches neglect the change of basis vectors in curved spaces, leading to an asymmetric joint-space stiffness matrix that violates passivity and conservation principles. In this work, we derive a physically consistent, symmetric joint-space stiffness formulation directly from the task-space stiffness matrix by explicitly incorporating Christoffel symbols. This correction resolves long-standing inconsistencies in stiffness modeling, ensuring energy conservation and stability. We validate our approach experimentally on a robotic system, demonstrating that omitting these correction terms results in significant asymmetric stiffness errors. Our findings bridge theoretical insights with practical control applications, offering a robust framework for stable and interpretable robotic interactions.
\end{abstract}


\maketitle

\section{Introduction}
In the realm of robotics, the concept of ``controller design in the physical domain'' \citep{Sharon_1989, Sharon_1991} and the associated methodology of ``control by interconnection'' \citep{stramigioli_modeling_2001, vanDerSchaft_2016} emphasize that robot controllers should be more than mere signal processors. Instead, they should have a physical interpretation \citep{lachner2022geometric}, which is especially important for robots that physically interact with the environment \citep{hogan_stability_1988, Dietrich2022, Hogan_2022}. This paper delves into impedance control \citep{hogan_impedance_1984} during physical interaction, specifically focusing on the symmetry of the stiffness matrix and its role in ensuring passive physical equivalent robot controllers.

Passivity is a fundamental property for ensuring coupled stability when interacting with arbitrary passive objects \citep{colgate_robust_1988}. Stability in robotics can be achieved by monitoring and controlling the energy supplied by the controller \citep{Colgate_1987, Colgate_1987_2, colgate_robust_1988, stramigioli_energy-aware_2015}. In impedance-controlled robots, this monitoring is particularly straightforward, as energy is stored in virtual springs (potential energy) and transferred into kinetic energy during movement \citep{lachner_energy_2021}.

During physical interaction, stiffness plays a crucial role, as it defines how energy is stored and exchanged between the robot and its environment. Task-space stiffness determines interaction forces due to contact, which is especially important at low frequencies (e.g., steady-state). Ensuring symmetry of the stiffness matrix in impedance-controlled robots is crucial for passive physical equivalent control, as an asymmetric stiffness matrix implies non-conservative force fields that contradict fundamental physical principles and violate passivity.

Prior studies have identified task-space stiffness asymmetry in robotic control \citep{CiblakLipkin_1994, Zefran1996, Zefran_1997, howard_CartK_1998, zefran_CartK_1998, zefran_metrics_1999, Chen_2000, Woolfrey2024}. This work builds upon these insights to present a physically consistent resolution that guarantees symmetric stiffness.

In this paper, we demonstrate how to derive a symmetric stiffness matrix in joint space coordinates. By explicitly accounting for the robot’s kinematic changes under external forces, we identify two asymmetric components whose sum results in a symmetric stiffness. To our knowledge, this is the first work to derive a physically consistent symmetric kinematic stiffness, incorporating correction terms via Christoffel symbols. Unlike previous studies concluding with an asymmetric stiffness matrix, our approach ensures energy conservation and can be effectively used for robot control during physical interactions. To validate our approach, we implement our method on a real robot and provide open-source code for users to apply our findings.

\section{Methods}\label{sec:Methods}
In this paper, we employ tensor notation and the Einstein summation convention to explicitly consider basis vectors (appendix~\ref{ap:Einstein}). All notations related to twists $\xi \in se(3)$ and wrenches $F \in se(3)$ can be found in appendix~\ref{ap:Twists}.

\subsection{Affine connection and Christoffel Symbols}\label{subsec:CS}
An affine connection provides a systematic way to compare vectors at different points on a manifold, defining how basis vectors evolve along coordinate directions and enabling differentiation in curved spaces. The Christoffel symbols $\Gamma^k_{\ ij}$ are the connection coefficients associated with an affine connection, capturing how basis vectors change as they move along the manifold.

A connection can be symmetric or asymmetric, depending on whether it has torsion. In a symmetric connection, the Christoffel symbols satisfy $\Gamma^k_{\ ij} = \Gamma^k_{\ ji}$, meaning the basis change is independent of the order of differentiation. In contrast, an asymmetric connection introduces torsion, leading to Christoffel symbols that do not necessarily exhibit this symmetry.

\begin{figure}[H]
\centering
\includegraphics[trim={0.25cm 0.0cm 0cm 0.0cm}]{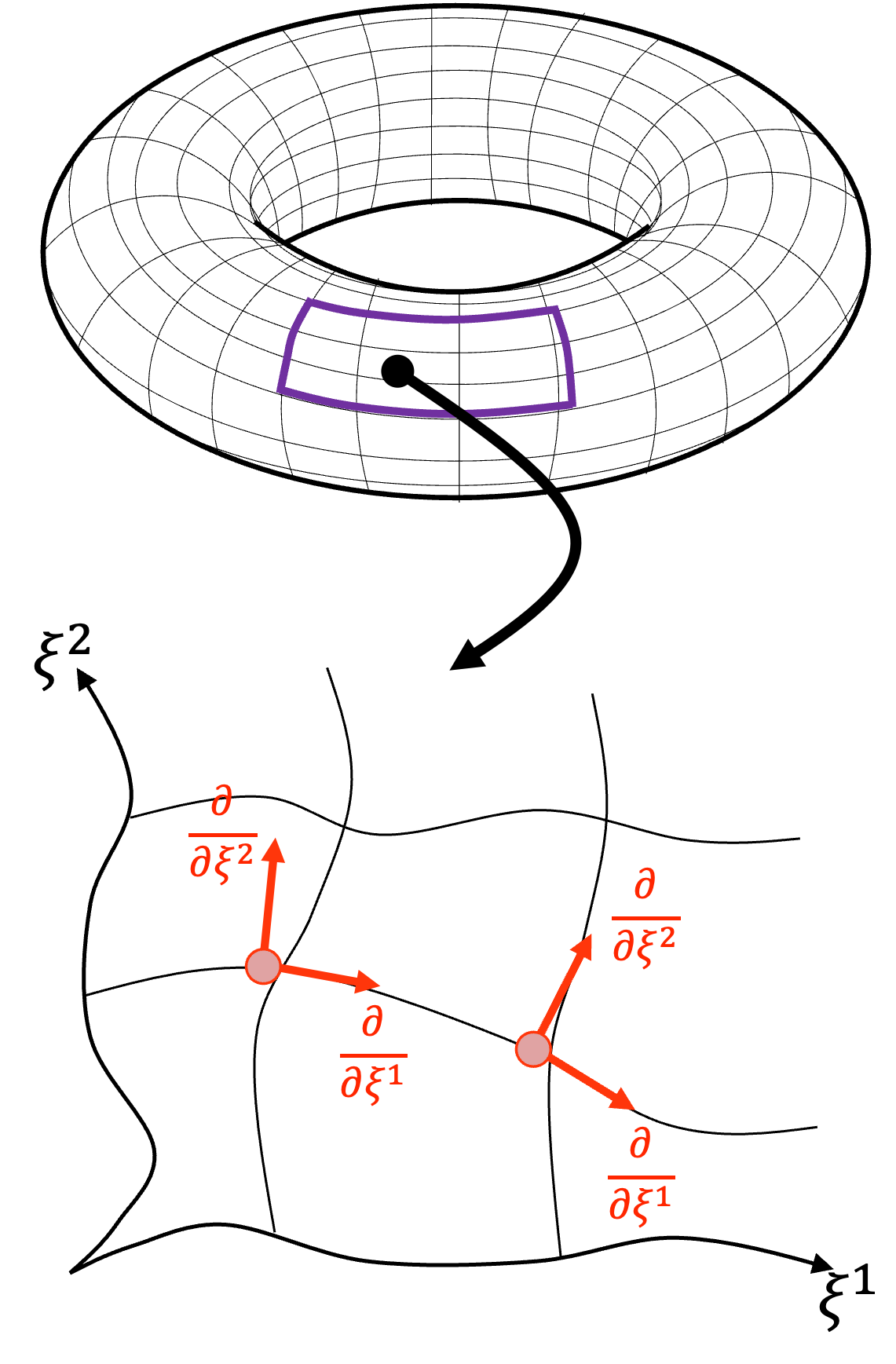}
\caption{Chart map of a toroidal manifold patch (highlighted in purple). The coordinate-induced basis vectors $\frac{\partial}{\partial \xi^1}$ and $\frac{\partial}{\partial \xi^2}$ at different locations illustrate the rotation of the basis due to the manifold’s curvature. This necessitates correction terms (Christoffel symbols) to properly account for the basis change when computing derivatives.}
\label{fig:CS_mfd}
\end{figure}

To perform differentiation on curved manifolds, it is essential to account for how the basis vectors $e_i$ evolve along coordinate directions $\xi ^j$. The vectors $e_i$ represent basis vector fields of the tangent space, which describe unit directions at different points on the manifold. In contrast, the coordinate-induced basis vectors $\frac{\partial}{\partial \xi^j}$ arise from the coordinate chart and describe differentiation along coordinate lines. The Christoffel symbols quantify how the basis vectors change due to curvature, ensuring that the differentiation aligns with the underlying geometry (fig.~\ref{fig:CS_mfd}).

The dependence of basis vectors $e_i$ on the coordinate components $\xi^j$ leads to changes that can be expressed as:
\begin{equation}\label{eq:ChristoffelSymbols}
\frac{\partial e_i}{\partial \xi^j} = \Gamma^m_{\ \ ij} \ e_m.
\end{equation}

Eq.~\eqref{eq:ChristoffelSymbols} shows that the Christoffel symbols $\Gamma^m_{\ \ ij}$  describe the rate at which the basis vector $e_i$ changes as we move along the coordinate $\xi^j$. This change is expressed in terms of the basis vector $e_m$, effectively measuring how the basis itself changes due to the curvature of the space. 

In a Cartesian space with zero curvature, the Christoffel symbols vanish, i.e., $\Gamma^m_{\ \ ij} = 0$. This is the case for pure translations, which are free vectors—shifting the origin does not alter their intrinsic properties, such as direction and magnitude. 

However, in curved spaces, the Christoffel symbols become essential, as the basis vectors themselves change with position. A key example is spatial rotations, which belong to the Lie group $SO(3)$, a curved and nonlinear manifold. Unlike Cartesian vectors, rotations do not follow standard vector operations: vector addition is not defined, composition is non-commutative ($R_1 R_2 \neq R_2 R_1$). These same principles extend to directional derivatives, where the order of differentiation matters due to the underlying geometry.

This leads to a fundamental property of the Christoffel symbols, which follows from the non-commutativity of directional derivatives:

\begin{equation*}
\frac{\partial e_i}{\partial \xi^j} - \frac{\partial e_j}{\partial \xi^i} = C^m_{\ \ ij} \ e_m.
\end{equation*}

Here, the structure coefficients $C^m_{\ \ ij}$ quantify the non-commutativity of the coordinate-induced basis vectors. In the context of robotic control, these coefficients correspond to the Lie algebra $se(3)$, whose explicit form is provided in Appendix~\ref{ap:StructureConstSE3}.

This directly implies that the Christoffel symbols satisfy:
\begin{equation}\label{eq:StructureCoef_CS}
\Gamma^m_{\ \ ij} - \Gamma^m_{\ \ ji} = C^m_{\ \ ij}.
\end{equation}

Thus, the presence of nonzero structure coefficients $C^m_{\ \ ij}$ introduces an asymmetry in the Christoffel symbols, reflecting the geometric structure of the coordinate basis. This asymmetry influences how differential changes in position and orientation affect the local basis vectors, which in turn influences how twists and wrenches are transformed under coordinate changes in task-space.

Since stiffness is defined as the rate of change of a wrench with respect to spatial displacement, it is inherently tied to how wrenches transform under coordinate variations. As we will see in the next section, the Christoffel symbols appear explicitly in the stiffness formulation.

In this paper, we adopt the \textit{Kinematic Connection}, originally introduced by \cite{zefran_metrics_1999} for body-fixed coordinates (also referred to as the left-invariant connection). A detailed derivation of this connection is presented in appendix~\ref{ap:KinConnection}.

\subsection{Task-space stiffness}
The task-space stiffness matrix $K_{ij}$ characterizes how wrenches $F_i$ change in response to spatial displacements $\xi^j$. It is defined as:
\begin{equation}\label{eq:Stiffness}
    K_{ij} = \frac{\partial F_i}{\partial \xi^j}. 
\end{equation}

\begin{figure}[H]
    \centering
    \includegraphics[width=0.7\textwidth]{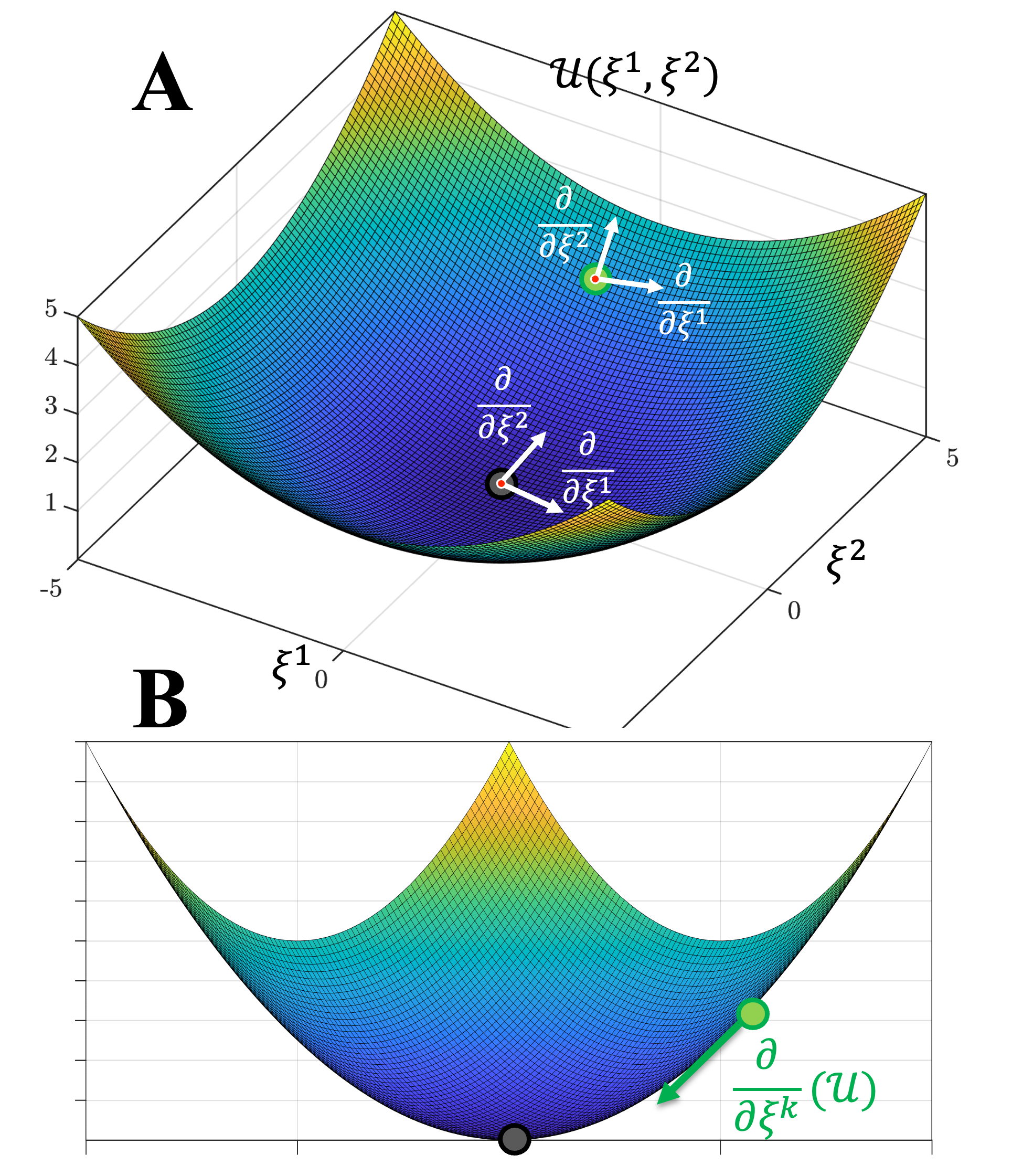}
    \caption{(A) Quadratic potential function $\mathcal{U}(\xi^1, \xi^2)$  with coordinate-induced basis vectors $\frac{\partial}{\partial \xi^1}$ and  $\frac{\partial}{\partial \xi^2}$ at different locations. The global minimum is marked by the black point. (B) At the green point, the presence of an external force (negative gradient $\frac{\partial}{\partial \xi^k}(\mathcal{U})$) induces a basis shift and rotation. This basis change, due to the curvature of the potential field, requires correction terms (Christoffel symbols).}
    \label{fig:ChangeBasis}
\end{figure}

The wrench components $F_i$ can be defined from the gradient of the potential energy function $\mathcal{U}$ (fig.~\ref{fig:ChangeBasis}B):
\begin{equation}
    F_i = \frac{\partial \mathcal{U}}{\partial \xi^j}
\end{equation}

In coordinate-free notation, the wrench can be described as the basis vector field $e_i$, acting as a directional derivative: 
\begin{equation}\label{eq:DirectDerivative}
    F_i = e_i(\mathcal{U})
\end{equation}
Here, the notation $e_i(\mathcal{U})$ denotes the application of the directional derivative $e_i$ to the scalar field $\mathcal{U}$, rather than a product. 

Substituting the expression for $F_i$ from eq.~\eqref{eq:DirectDerivative} into eq.~\eqref{eq:Stiffness}, we obtain:
\begin{equation}
K_{ij} = \frac{\partial }{\partial \xi^j} \Big( e_i(\mathcal{U}) \Big).
\end{equation}

Applying the Leibniz rule for differentiation, we expand the derivative as:
\begin{equation}\label{eq:Stiffness_Leibniz}
    K_{ij} = \Big( \frac{\partial e_i}{\partial \xi^j} \Big) (\mathcal{U}) + e_i \Big( \frac{\partial \mathcal{U}}{\partial \xi^j} \Big).
\end{equation}
The first term, $\big( \frac{\partial e_i}{\partial \xi^j} \big) (\mathcal{U})$, accounts for the variation of the basis vector field $e_i$ with respect to coordinate $\xi^j$. The second term, $e_i \big( \frac{\partial \mathcal{U}}{\partial \xi^j} \big)$, represents the directional derivative of the gradient of $\mathcal{U}$ along $e_i$.

This formulation explicitly considers changes in the basis vectors, ensuring a physically meaningful representation of stiffness in curved spaces.

This step is crucial in the derivation, as the term $\frac{\partial e_i}{\partial \xi^j}$ explicitly captures how the basis vectors $e_i$ must be adjusted when shifted along the curved coordinate components $\xi^j$. An illustrative example is shown in fig.\ref{fig:ChangeBasis}: an external force prevents the system from reaching the global minimum of the potential energy function. Instead, due to the curvature of the potential energy landscape, the local coordinate-induced basis is rotated relative to the global minimum. This correction is systematically encoded by the Christoffel symbols (sec.\ref{subsec:CS}). While conventional matrix formulations neglect changes in the basis, tensor notation explicitly accounts for them, ensuring a complete representation of the system’s behavior.

From the definition of the Christoffel Symbols (eq.~\eqref{eq:ChristoffelSymbols}), we know that:
\begin{equation}
\frac{\partial e_i}{\partial \xi^j} = \Gamma^m_{\ \ ij} \ \frac{\partial}{\partial \xi^m}.
\end{equation}
Using the coordinate-induced basis representation $e_i = \frac{\partial}{\partial \xi^i}$, we rewrite eq.\eqref{eq:Stiffness_Leibniz} as:
\begin{equation}
    K_{ij} = \frac{\partial^2 \mathcal{U}}{\partial \xi^j \partial \xi^j} + \Gamma_{\ \ ij}^{m} \ \frac{\partial \mathcal{U}}{\partial \xi^m}.
\end{equation}

Since the wrench components are defined as $F_m = \frac{\partial \mathcal{U}}{\partial \xi^m}$, we arrive at the final expression:
\begin{equation}\label{eq:Kij_fin}
K_{ij} = \frac{\partial^2 \mathcal{U}}{\partial \xi^i \partial \xi^j} + \Gamma_{\ \ ij}^{m} \ F_m.
\end{equation}

As can be seen, the task-space stiffness matrix consists of two components: the second partial derivatives of the potential energy function and a correction terms, involving the external wrench $F_m$, which arises due to coordinates basis change.

\subsection{Problem Statement}
Conventional matrix formulations overlook the change of basis vectors, leading to the conclusion that the task-space stiffness matrix is solely given by the second partial derivatives of the potential energy function. This formulation inherently yields a symmetric stiffness matrix.

By explicitly accounting for basis changes, eq.~\eqref{eq:Kij_fin} introduces a correction term $\Gamma_{\ \ ij}^{m} F_m$ which depends on the Christoffel symbols. This means that the symmetry of the task-space stiffness $K_{ij}$ is directly influenced by the symmetry properties of the Christoffel symbols themselves.

The asymmetry of the task-space stiffness matrix in robotic control has been extensively reported in the literature \citep{CiblakLipkin_1994, Zefran1996, Zefran_1997, howard_CartK_1998, zefran_CartK_1998, zefran_metrics_1999, Chen_2000, Woolfrey2024}.

An asymmmetric stiffness is not usable for passive physical equivalent controllers, as it is equivalent to implementing a perpetual motion machine of the first kind, as will be shown in the following. 

The task-space stiffness $K_{ij}$ can be decomposed into symmetric and anti-symmetric (skew-symmetric) components:
\begin{equation}\label{eq:StiffnessDecomposition}
K_{ij} = \underbrace{\frac{1}{2} (K_{ij} + K_{ji})}_{\text{symmetric part}} + \underbrace{\frac{1}{2} (K_{ij} - K_{ji})}_{\text{antisymmetric part}}.
\end{equation}

The antisymmetric component evokes a force orthogonal to displacement . 
The (incremental) mechanical work done by a trajectory of displacements is $\Delta W = \int F_i \ \Delta x^i$.
If the displacement trajectory traverses a closed orbit of non-zero size, the system will exhibit a net energy gain or loss per cycle, leading to energy accumulation or dissipation. Traversing this orbit multiple times would implement a perpetual motion machine of the first kind. No physical spring can do this.

A robot-implemented virtual spring may be able to implement this behavior. However, this has a clear disadvantage: \textit{violation of passivity}.

In the next section, we derive a physically meaningful, symmetric stiffness matrix in joint-space coordinates, which is obtained from the task-space stiffness matrix. By explicitly incorporating Christoffel symbols, we demonstrate how to recover a physically consistent stiffness formulation that accounts for basis vector changes, adheres to conservation principles, and ensures passive control in contact-rich manipulation.

\section{Results}
In this section, Greek letters denote joint space coordinate components, while Latin indices represent task-space coordinate components, consistent with the notation in Sec.~\ref{sec:Methods}.

\subsection{Theory}
A joint-space stiffness matrix $\mathcal{K}_{\alpha \beta}$ is the gradient of joint torque $\tau_{\beta}$ with respect to joint displacement $q^{\alpha}$:
\begin{equation}\label{eq:JointSpaceStiffness}
    \mathcal{K}_{\alpha \beta} = \frac{\partial \tau_{\beta}}{\partial q^{\alpha}}.
\end{equation}

We can map the task-space wrench $F_k$ to torques by the Jacobian map $b^k_{\ \beta}$:
\begin{equation}\label{eq:JacobianMap}
    \tau_{\beta} = b^k_{\ \beta} \ F_k.
\end{equation}
Substituting eq.~\eqref{eq:JacobianMap} into eq.~\eqref{eq:JointSpaceStiffness} yields:
\begin{equation}
    \mathcal{K}_{\alpha \beta} = \frac{\partial}{\partial q^{\alpha}}( b^k_{\ \beta} \ F_k ).    
\end{equation}
Applying the Leibniz rule for differentiation, this expands to:
 \begin{equation}\label{eq:LeipnizRule}     
    \mathcal{K}_{\alpha \beta} = \frac{\partial b_{\ \beta}^{k}}{\partial q^\alpha} F_k + b_{\ 
 \beta}^{k} \frac{\partial F_k}{\partial q^\alpha}.
 \end{equation}
The term $\frac{\partial F_k}{\partial q^\alpha}$ can be expanded using the chain rule:
\begin{equation}\label{eq:ExpansionJacobian}
    \frac{\partial F_k}{\partial q^\alpha} = \frac{\partial F_k}{\partial \xi^l} \frac{\partial \xi^l}{\partial q^\alpha} = \frac{\partial F_k}{\partial \xi^l} \ a_{\ \alpha}^{l},
\end{equation}
where $a_{\ \alpha}^{l}$ represents the Jacobian mapping between the coordinate systems. Substituting eq.\eqref{eq:ExpansionJacobian} into eq.\eqref{eq:LeipnizRule} yields:
\begin{equation}\label{eq:ExpandedLeipnizRule}
    K_{\alpha\beta} = \frac{\partial b_{\ \beta}^{k}}{\partial q^{\ \alpha}} F_k + b_{\ \beta}^{k} \ a_{\ \alpha}^{l} \frac{\partial F_k}{\partial \xi^l}.
\end{equation}

The wrench component $F_k$ is the partial derivative of the potential function $\mathcal{U}$ with respect to coordinate $\xi^k$, representing the gradient in the $k$-th direction. In coordinate-free notation, the basis vector $e_k$ of the tangent space acts as a directional derivative, expressed as $e_k (\mathcal{U})$: 
\begin{equation}\label{eq:Gradient}
    F_k = \frac{\partial \mathcal{U}}{\partial \xi^k} = e_k (\mathcal{U}).
\end{equation}

Substituting the expression for $F_k$ from eq.\eqref{eq:Gradient} into eq.\eqref{eq:ExpandedLeipnizRule}, we obtain:
\begin{equation}
K_{\alpha\beta} = \frac{\partial b_{\ \beta}^{k}}{\partial q^{\ \alpha}} F_k + b_{\ \beta}^{k} \ a_{\ \alpha}^{l} \frac{\partial}{\partial \xi^l} \left( e_k (\mathcal{U}) \right),
\end{equation}
where $e_k$ is a vector field and $\mathcal{U}$ is a function. The differentiation of their product follows the Leibniz rule and we expand the derivative as:
\begin{equation}\label{eq:ProductVecFieldFunction}
K_{\alpha\beta} = \frac{\partial b_{\ \beta}^{k}}{\partial q^{\alpha}} F_k
+ b_{\ \beta}^{k} \ a_{\ \alpha}^{l} \left( \frac{\partial e_k}{\partial \xi^l} (\mathcal{U})
+ e_k \frac{\partial \mathcal{U}}{\partial \xi^l} \right).
\end{equation}

Eq.~\eqref{eq:ProductVecFieldFunction} represents a key step in the derivation, as it explicitly accounts for the correction of basis vectors $e_k$ along the curved coordinate components $\xi^l$ (fig.~\ref{fig:ChangeBasis}). This aspect is overlooked in conventional matrix formulations \citep{mussa1985neural, hogan_mechanics_1985, mussa-ivaldi_integrable_1991}. 

For a basis vectors, we have $e_k = \frac{\partial}{\partial \xi^k}$. Additionally, from eq.~\eqref{eq:ChristoffelSymbols}, we know that the basis derivatives satisfy:
\begin{equation}
\frac{\partial e_k}{\partial \xi^l} = \Gamma^m_{\ \ kl} \frac{\partial}{\partial \xi^m},
\end{equation}
which allows us to rewrite eq.\eqref{eq:ProductVecFieldFunction} as:
\begin{equation}
K_{\alpha\beta} = \frac{\partial b_{\ \beta}^{k}}{\partial q^{\alpha}} F_k
+ b_{\ \beta}^{k} \ a_{\ \alpha}^{l} \left( \Gamma_{\ \ kl}^{m} \ \frac{\partial \mathcal{U}}{\partial \xi^m}
+ \frac{\partial^2 \mathcal{U}}{\partial \xi^k \partial \xi^l} \right).
\end{equation}

Since $\frac{\partial \mathcal{U}}{\partial \xi^m}$ corresponds to the wrench component $F_m$ and $\frac{\partial^2 \mathcal{U}}{\partial \xi^k \partial \xi^l}$ represents the task-space stiffness matrix $K_{kl}$, we arrive at:
\begin{equation}\label{eq:JointSpaceStiffnessCorrection}
K_{\alpha\beta} = \frac{\partial b_{\ \beta}^{k}}{\partial q^{\alpha}} F_k
+ b_{\ \beta}^{k} \ a_{\ \alpha}^{l} \ \Gamma_{\ \ kl}^{m} \ F_m + b_{\ \beta}^{k} \ a_{\ \alpha}^{l} \ K_{kl}.
\end{equation}

This is the central finding of our paper. The term $(b_{\ \beta}^{k} \ a_{\ \alpha}^{l} \ K_{kl})$ is inherently symmetric, as it derives from the Hessian matrix, which possesses this property. In contrast, the kinematic stiffness $(\frac{\partial b_{\ \beta}^{k}}{\partial q^{\alpha}} F_k)$ can be asymmetric when rotations (i.e., moment components in $F_k$) are involved. An example can be found in appendix~\ref{ap:AntiSim_Kkin}. Crucially, we establish that their combination with the Christoffel term,
\begin{equation}
 \frac{\partial b_{\ \beta}^{k}}{\partial q^{\alpha}} F_k + b_{\ \beta}^{k} \ a_{\ \alpha}^{l} \ \Gamma_{\ \ kl}^{m} \ F_m,   
\end{equation}
is symmetric since we provide a connection that is consistent with the robot kinematics (see example in appendix~\ref{ap:Sim_Kkin}).

This result is significant because it demonstrates that the inclusion of Christoffel symbols ensures symmetry in the joint-space stiffness matrix, a property essential for stable contact-rich manipulation. While \cite{zefran_metrics_1999} showed that the task-space stiffness in general asymmetric, our approach corrects this, ensuring a physically equivalent control framework that adheres to fundamental principles like energy conservation.

While this derivation is theoretical, the next section presents experimental validation, highlighting its direct impact on real-world robotic interaction.

\subsection{Practical Experiments}
The experiments were conducted on a KUKA LBR iiwa with seven DOFs, using KUKA’s Fast Robot Interface (FRI) for torque control. Built-in gravity and Coriolis/centrifugal compensation remained active throughout. Kinematic and dynamic matrices were computed via the Exp[licit]-FRI interface\footnote{\url{https://github.com/explicit-robotics/Explicit-FRI}}. To follow standard robot control conventions, we present matrix notation in this subsection.

We implemented a joint-space impedance controller, where the control torque $\bm{\tau} \in \mathbb{R}^n$ was determined by joint-space stiffness $\bm{\mathcal{K}} \in \mathbb{R}^{n \times n}$, acting on the deviation between the equilibrium configuration $\bm{q}_0$ and the joint configuration $\bm{q}$, and joint-space damping $\bm{\mathcal{B}} \in \mathbb{R}^{n \times n}$, acting on the joint velocity $\dot{\bm{q}} \in \mathbb{R}^n$: 
\begin{equation}
    \bm{ \tau } = \bm{\mathcal{K}} (\bm{q}_0 - \bm{q} ) - \bm{\mathcal{B}} \ \dot{ \bm{q} }.
\end{equation}

\begin{figure}[H]
    \centering
\includegraphics[trim={0cm 0.0cm 0cm 0.0cm}, clip]{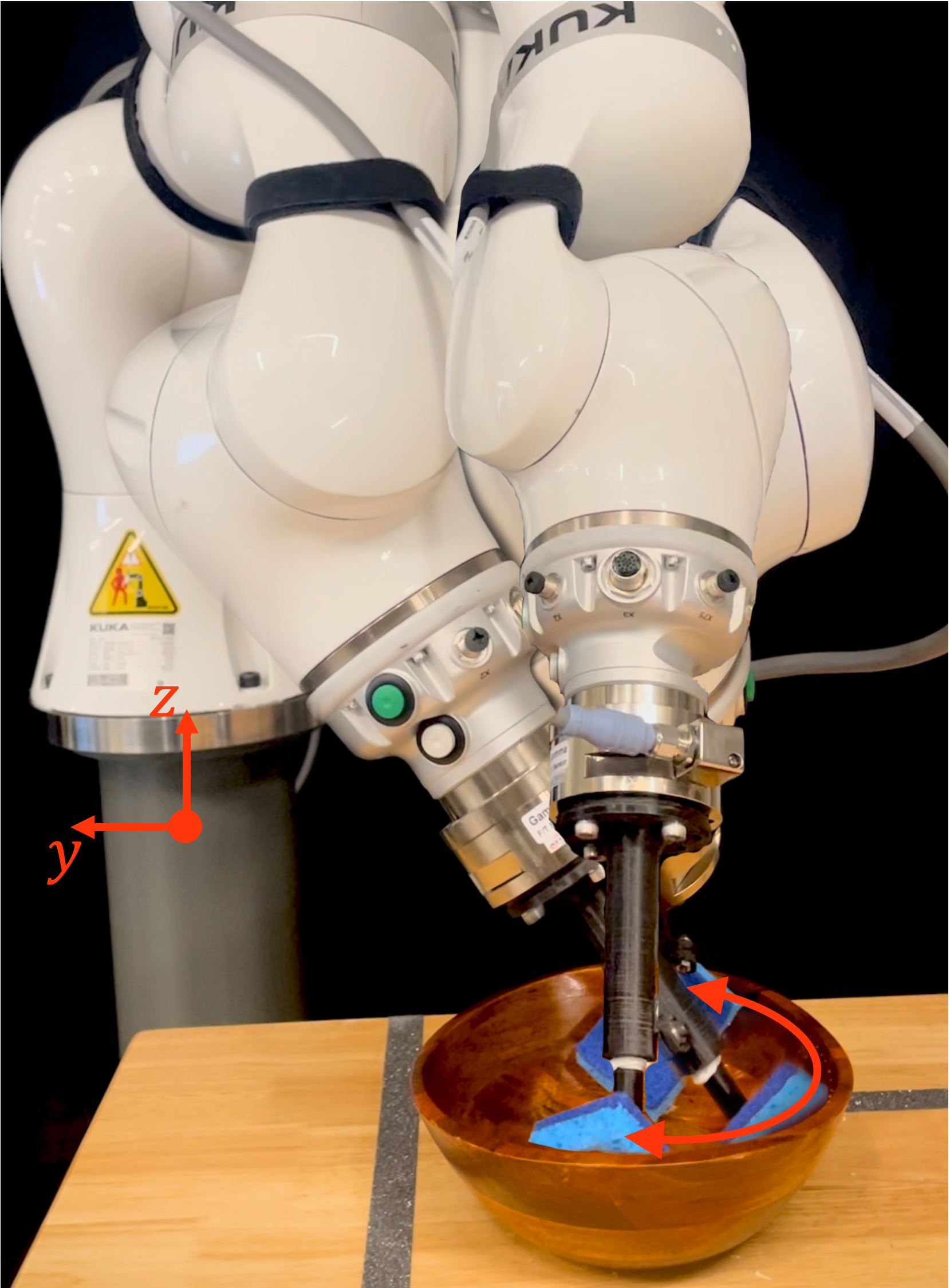}
  \caption{Robot movement, visualized as overlaid robot configurations and directions during the bowl-wiping task.}
  \label{fig:Experiment}
\end{figure}
The joint-space stiffness was computed using eq.~\eqref{eq:JointSpaceStiffnessCorrection}, incorporating the Jacobian matrix $\bm{J}(\bm{q}) \in \mathbb{R}^{6 \times 7}$, task-space stiffness $\bm{K} \in \mathbb{R}^{6 \times 6}$, and correction terms from Christoffel symbols and external wrenches $(\bm{\Gamma} \bm{F}_{\text{ext}}) \in \mathbb{R}^{6 \times 6}$ (appendix~\ref{ap:CorrectionTerms}). The partial derivatives of $\bm{J}(\bm{q})^T$ were derived symbolically using Exp[licit]-Matlab\footnote{\url{https://github.com/explicit-robotics/Explicit-MATLAB}} \citep{lachner2024explicit} and integrated into the real-time code, available on the paper’s GitHub repository\footnote{\url{https://github.com/jlachner/Physically-Consistent-Stiffness-Formulation.git}}. In matrix form, eq.~\eqref{eq:JointSpaceStiffnessCorrection} can be written as:
\begin{equation}\label{eq:KinStiffMatrixForm}
\bm{\mathcal{K}} = \frac{ \partial \bm{J}(\bm{q})^T }{ \partial \bm{q} } + \bm{J}(\bm{q})^T \ \left[ \bm{K} + \bm{\Gamma} \bm{F}_{\text{ext}} \right] \ \bm{J}(\bm{q}).
\end{equation}

The joint-space damping matrix $\bm{\mathcal{B}}$ was designed based on $\bm{\mathcal{K}}$ and the robot’s inertia matrix $\bm{\mathcal{M}}(\bm{q}) \in \mathbb{R}^{7 \times 7}$ \citep{albu-schaffer_cartesian_2003}.

The robot task was to wipe a bowl in a semi-circular motion (Fig.~\ref{fig:Experiment}) while exerting significant external wrenches, measured by an ATI Force-Torque sensor attached to the robot end-effector. The zero-force trajectory $\bm{q}_0$ was generated using Exp[licit]-Matlab. The code can be found on the paper's Github repository. 

Two experimental trials were conducted: one with correction terms ($\bm{\Gamma} \bm{F}_{\text{ext}}$ in eq.\eqref{eq:KinStiffMatrixForm}) and one without, as convention in conventional matrix formulations \citep{mussa1985neural, hogan_mechanics_1985, mussa-ivaldi_integrable_1991}. 

\begin{figure}[h]
    \centering
    \includegraphics[trim={0cm 0.0cm 0cm 0.0cm}, clip]{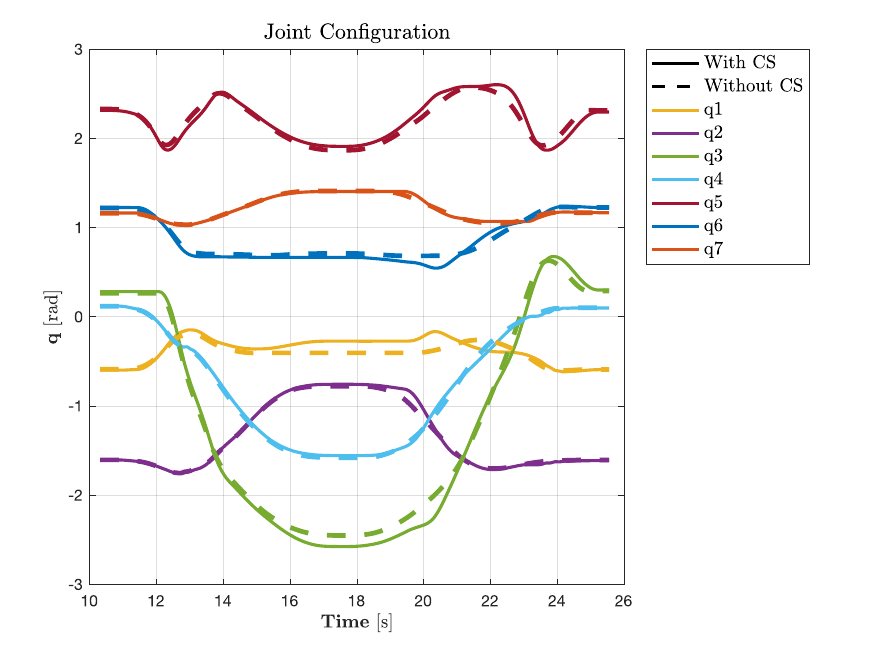}
    \caption{Recorded robot joint configurations during both
experimental trials. The solid line represents the trial with correction terms (Christoffel symbols), and the dashed line without.}
    \label{fig:RobotConfigs}
\end{figure}

As can be seen in fig.~\ref{fig:RobotConfigs}, the correction terms had an influence on the robot trajectory, compared to the standard approach.

\begin{figure}[h]
    \centering
    \includegraphics[width=1\textwidth,trim={0cm 0.0cm 0cm 0.0cm}, clip]{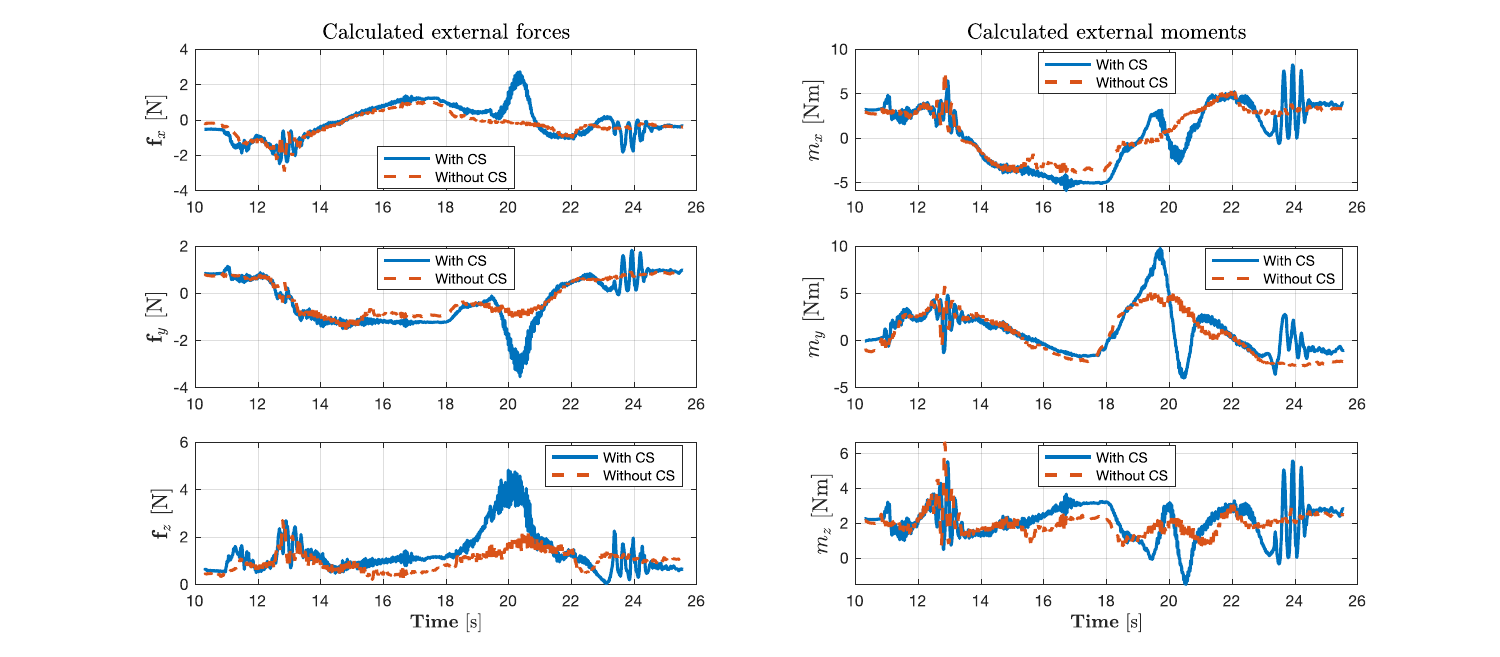}
    \caption{External forces and moments from the ATI force-torque sensor, expressed in the base frame. The solid line represents the trial with correction terms (Christoffel symbols), and the dashed line without.}
    \label{fig:GenForces}
\end{figure}

The recorded external forces and moments during both experimental trials can be seen in fig.~\ref{fig:GenForces}.  
The applied moments at the end-effector, ranging between -5 Nm and 10 Nm, affected the symmetry of the joint-space stiffness matrix. To quantify the asymmetry, the stiffness matrix was decomposed into symmetric and anti-symmetric components (eq.~\eqref{eq:StiffnessDecomposition}).

\begin{figure}[h]
    \centering
    \includegraphics[width=1\textwidth,trim={0cm 0.0cm 0cm 0.0cm}, clip]{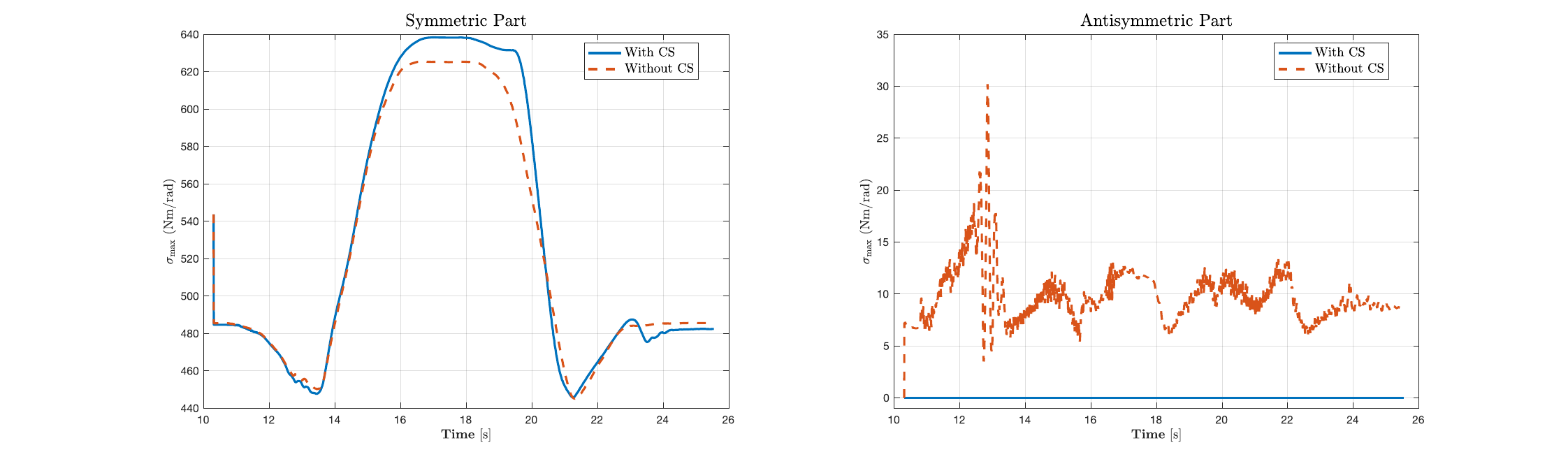}    
    \caption{Maximum singular values of the symmetric (left) and anti-symmetric (right) parts of the joint-space stiffness matrix during the experiments. The solid line represents the trial with correction terms (Christoffel symbols), and the dashed line without.}
  \label{fig:MaxSingK}
\end{figure}

Fig.~\ref{fig:MaxSingK} shows the maximum singular values of both components over time.
The results confirm that the correction term ensured the symmetry of the joint-space stiffness matrix (cf. zero anti-symmetric component in fig.~\ref{fig:MaxSingK}). In contrast, omitting the correction term led to an anti-symmetric part with singular values reaching approximately 30 Nm/rad around 13 seconds.

This stiffness error is significant. For comparison, the torque required to tighten a standard M10 bolt with a hand wrench is about 20 Nm. The observed stiffness error corresponds to one and a half times that effort, highlighting the practical impact of neglecting the correction terms.

\section{Discussion}
This work demonstrates that ensuring the symmetry of the stiffness matrix is essential for passive physically equivalent control in contact-rich manipulation. Conventional matrix formulations neglect the change of basis vectors in curved spaces, leading to an asymmetric kinematic stiffness matrix in joint-space coordinates. By explicitly incorporating Christoffel symbols, we derived a symmetric stiffness formulation that adheres to conservation principles, ensuring passivity and stability.

The proposed approach resolves long-standing inconsistencies in task-space stiffness modeling, where prior work acknowledged asymmetries but lacked a physically consistent correction. Our experimental validation on a real robotic system confirms that including correction terms significantly impacts the system’s response, ensuring that the joint-space stiffness remains symmetric under external wrenches.

While the method guarantees physical consistency, its practical implementation requires additional computational resources due to the Christoffel term calculations. Future work may focus on optimizing these computations for real-time control and extending the approach to adaptive or learning-based frameworks. Additionally, integrating this formulation into whole-body control strategies could further enhance robotic interaction capabilities.

By bridging theoretical foundations with practical implementation, our work provides a robust framework for impedance control that respects fundamental physical principles, paving the way for stable robotic interactions.

\section{Conclusion}
This paper presents a passive physically equivalent approach for contact-rich manipulation, resolving a long-standing issue of asymmetric stiffness in impedance-controlled robots. By explicitly accounting for the geometric properties of robot kinematics, we derived a symmetric stiffness formulation in joint-space coordinates, incorporating correction terms derived from Christoffel symbols. Unlike previous formulations that resulted in inherently asymmetric stiffness matrices, our method guarantees compliance with fundamental physical principles, including passivity and energy conservation.

We demonstrated that the presence of asymmetric stiffness terms leads to non-conservative force fields, violating passivity and introducing artificial energy sources into the system. Through theoretical analysis and experimental validation, we showed that including Christoffel symbol-based corrections ensures a symmetric stiffness matrix, preventing unphysical energy accumulation and enabling stable interaction forces in real-world robotic tasks.

Our experimental results on a KUKA LBR iiwa robot confirmed that neglecting these correction terms can lead to substantial stiffness errors, potentially compromising the stability of impedance-controlled robots in contact-rich environments. By integrating the proposed corrections, we restored the expected physical behavior, yielding stable and interpretable robotic interactions.

\section*{Acknowledgements}
We gratefully acknowledge the support of KUKA, an international leader in automation solutions, and specifically thank them for providing the KUKA robots used in our experiments.

The authors would like to thank Dr. Federico Tessari for his assistance in setting up the experimental setup.

\subsection*{Funding}
This work was supported in part by the MIT/SUSTech Centers for Mechanical Engineering Research and Education. 
MCN was supported in part by a Mathworks Fellowship. 
JL was supported in part by the MIT-Novo Nordisk Artificial Intelligence Postdoctoral Fellows Program.

\subsection*{Declaration of Conflicting Interests}
The Authors declare that there is no conflict of interest.


\bibliographystyle{SageH}
\bibliography{literature}

\begin{appendices}

\section{Tensor geometry}\label{ap:Einstein}

Consider a point $\bm{q}$ on a $n$-dimensional differentiable manifold $Q$. For all curves on this manifold going through point $\bm{q}$, a vector space $\mathcal{V}$ exists that collects all \textit{contravariant elements} tangent to the curves. In the point $\bm{q}$ also a dual vector space $\mathcal{V}^*$ exists, collecting all \textit{covariant elements}.
\subsection{Covariant tensors}
A covariant tensor is a multilinear function of type:
\begin{equation}\label{covariantTensor}
\bm{Q}: \underbrace{\mathcal{V} \times ... \times \mathcal{V}}_\textit{s many} \rightarrow \mathbb{R}.
\end{equation}
The function $\bm{Q}$ takes $s$ many elements of $\mathcal{V}$ and returns a real  number. A covariant tensor with $s = 1$ is called a co-vector. For example, a wrench is represented with one covariant subscript $i$: $F_{i}$.
\subsection{Contravariant tensors}
A contravariant tensor is a multilinear function on a covariant tensor:
\begin{equation}\label{contravariantTensor}
\bm{P}: \underbrace{\mathcal{V}^{*} \times ... \times \mathcal{V}^{*}}_\textit{r many} \rightarrow \mathbb{R}.
\end{equation}
The function $\bm{P}$ takes $r$ many elements of $\mathcal{V}^{*}$ and returns a real  number. A contravariant tensor with $r = 1$ is called a vector. For example, a twist is represented with one contravariant superscript $i$: $\xi^{i}$.
\subsection{Mixed tensors}
Mixed tensors are $r$-times contravariant and $s$-times covariant:
\begin{equation}\label{tensorDef}
\bm{T}: \underbrace{\mathcal{V}^{*} \times ... \times \mathcal{V}^{*}}_\textit{r many} \times \underbrace{\mathcal{V} \times ... \times \mathcal{V}}_\textit{s many} \rightarrow \mathbb{R}.
\end{equation}
For function $\bm{T}$, the notation $\begin{pmatrix}r\\s\end{pmatrix}$ can be used. The number $(r+s)$ defines the rank of the tensor. The tensor can be denoted with $r$ many contravariant superscripts and $s$ many covariant subscripts. For example, a $\begin{pmatrix}1\\1\end{pmatrix}$-tensor with rank 2 has one contravariant superscript $i$ and one covariant subscript $j$: $T^{i}_{\ j}$. A quadratic form (e.g., inertia) is represented by two covariant subscripts $i$ and $j$: $g_{ij}$

\subsection{Tensor contraction and Einstein summation convention}
Operations on tensors can be illustrated with an example:
\begin{equation}\label{tensorExample}
    \eta_k = \sum_{i=1}^n g_{ik} \ \xi^i,
\end{equation}
with $i, k = 1,...,n$. On the right side of eq.~\eqref{tensorExample}, a contravariant tensor $\xi^i$ (i.e., a vector) and a twice covariant tensor $g_{ik}$ appears (i.e., a metric tensor, which serves as the kernel of a quadratic form). The left side yields a covariant tensor $\eta_k$ (i.e., a co-vector). The operation on the appearing tensors can be seen: The tensor $g_{ik}$ is used to transform the contravariant tensor $\xi^i$ to a covariant tensor $\eta_k$. This operation on tensors is called ``contraction.'' Contraction sums pairs of equal indices, which are one upper (contravariant) index and one lower (covariant) index. As a result, the index can be canceled out. Implicitly, the sum symbol is left out, which is called ``Einstein summation convention'': 
\begin{equation}
    \eta_k = g_{ik} \ \xi^i.
\end{equation}
More details about tensors can be found in \cite{dubrovin_modern_1984}.

\section{Twists and Wrenches}\label{ap:Twists}
The set of rigid body transformations $\bm{H}$ form the manifold $SE(3)$. $SE(3)$ is not only a smooth manifold but also a group and therefore is called a ``Lie group.'' The tangent space to the identity element $\mathcal{I}$ of the group is denoted $T_\mathcal{I}SE(3)$. For $SE(3)$, the identity element is the $4 \times 4$ identity matrix. Elements of $T_{\bm{q}}Q$ and $T_\mathcal{I}SE(3)$ are called ``tangent vectors.'' 

One can find a local coordinate representation of tangent vectors with respect to the coordinate-induced bases. Tangent vectors of $T_\mathcal{I}SE(3)$ are called ``twists.'' A twist can either be an infinitesimal displacement or a velocity. It can be expressed in a fixed inertial coordinate frame (called ``spatial twist'') or in a moving body-fixed coordinate frame (called ``body twist''). The twist $\bm{\Tilde{\xi}} \in se(3)$ is calculated by $\dot{\bm{H}}{\bm{H}}^{-1}$ for spatial coordinates and calculated by ${\bm{H}}^{-1}\dot{\bm{H}}$ for body coordinates, with respect to basis $\Tilde{e}_i$ (matrix form).

Tangent vectors of $T_\mathcal{I}SE(3)$ are locally isomorphic to $\mathbb{R}^6$. If $T_\mathcal{I}SE(3)$ is associated with $\mathbb{R}^6$, the 0-element of the Lie algebra is the zero vector. The twists $\dot{\bm{H}}{\bm{H}}^{-1}$ and ${\bm{H}}^{-1}\dot{\bm{H}}$ can be represented as six-dimensional vectors. The basis are denoted $e_i$ (vector form). For the representation in this paper, the coordinate components of the twist represent the linear displacement or velocity $\dot{\bm{p}} = (\xi_1, \xi_2, \xi_3)^T$ and angular velocity $\bm{\omega} = (\xi_4, \xi_5, \xi_6)^T$.

\begin{equation*}
    \bm{\Tilde{\xi}} = \xi^i \ \Tilde{e}_i.
\end{equation*}
Here, $\xi^i$ are the vector components and $\Tilde{e}_i$ are the standard basis of $se(3)$:
\newline
\begin{center}
$\Tilde{e}_1 = \begin{pmatrix} 
0, 0, 0, 1 \\
0, 0, 0, 0 \\
0, 0, 0, 0 \\
0, 0, 0, 0
\end{pmatrix},
\Tilde{e}_2 = \begin{pmatrix} 
0, 0, 0, 0 \\
0, 0, 0, 1 \\
0, 0, 0, 0 \\
0, 0, 0, 0
\end{pmatrix}, \newline
\Tilde{e}_3 = \begin{pmatrix} 
0, 0, 0, 0 \\
0, 0, 0, 0 \\
0, 0, 0, 1 \\
0, 0, 0, 0
\end{pmatrix},
\Tilde{e}_4 = \begin{pmatrix} 
0, 0, 0, 0 \\
0, 0, -1, 0 \\
0, 1, 0, 0 \\
0, 0, 0, 0
\end{pmatrix}, \newline
\Tilde{e}_5 = \begin{pmatrix} 
0, 0, 1, 0 \\
0, 0, 0, 0 \\
-1, 0, 0, 0 \\
0, 0, 0, 0
\end{pmatrix},
\Tilde{e}_6 = \begin{pmatrix*}
0, -1, 0, 0 \\
1, 0, 0, 0 \\
0, 0, 0, 0 \\
0, 0, 0, 0
\end{pmatrix*}.$
\newline
\end{center}
Matrices $\Tilde{e}_1, \Tilde{e}_2, \Tilde{e}_3$ represent unit-directions for the linear motion along the $x,y,$ and $z-$coordinate and matrices $\Tilde{e}_4, \Tilde{e}_5, \Tilde{e}_6$ represent unit rotations about orthogonal axes.

The space of linear maps from $T_\mathcal{I}SE(3)$ to real numbers is called the ``co-tangent space'' $T^*_\mathcal{I}SE(3)$. $T^*_\mathcal{I}SE(3)$ has the structure of a Lie algebra, indicated as $se^*(3)$. Moreover, $T^*_\mathcal{I}SE(3)$ is a vector space which is called the ``co-vector space.'' A generalized force $\bm{F}$ is an element of the co-vector space and will be called a ``wrench.'' Like a twists, the wrench $\bm{F}$ can be represented with respect to dual basis $\Tilde{\lambda}^i$ (matrix form) and $\lambda^i$ (vector form). The dual basis is a set of co-vectors that satisfy the Kronecker delta $\delta^i_{\ j} = \lambda^i(e_j)$, which is $1$ for $i = j$ and $0$ for $i \neq j$. For the representation in this paper, the coordinate components of the wrench represent the linear force $\bm{f} = (F_1, F_2, F_3)^T$ and moment $\bm{m} = (F_4, F_5, F_6)^T$.

\begin{equation*}
    \bm{\Tilde{F}} = F_i \ \Tilde{\lambda}^i.
\end{equation*}
Here, $\Tilde{F}_i$ are the co-vector components and $\Tilde{\lambda}^i$ are the standard basis of $se^*(3)$:
\newline
\begin{center}
$\Tilde{\lambda}^1 = \begin{pmatrix} 
0, 0, 0, 1 \\
0, 0, 0, 0 \\
0, 0, 0, 0 \\
0, 0, 0, 0
\end{pmatrix},
\Tilde{\lambda}^2 = \begin{pmatrix} 
0, 0, 0, 0 \\
0, 0, 0, 1 \\
0, 0, 0, 0 \\
0, 0, 0, 0
\end{pmatrix}, \newline
\Tilde{\lambda}^3 = \begin{pmatrix} 
0, 0, 0, 0 \\
0, 0, 0, 0 \\
0, 0, 0, 1 \\
0, 0, 0, 0
\end{pmatrix},
\Tilde{\lambda}^4 = \begin{pmatrix} 
0, 0, 0, 0 \\
0, 0, -1, 0 \\
0, 1, 0, 0 \\
0, 0, 0, 0
\end{pmatrix}, \newline
\Tilde{\lambda}^5 = \begin{pmatrix} 
0, 0, 1, 0 \\
0, 0, 0, 0 \\
-1, 0, 0, 0 \\
0, 0, 0, 0
\end{pmatrix},
\Tilde{\lambda}^6 = \begin{pmatrix*}
0, -1, 0, 0 \\
1, 0, 0, 0 \\
0, 0, 0, 0 \\
0, 0, 0, 0
\end{pmatrix*}.$
\newline
\end{center}

Matrices $\Tilde{\lambda}^1, \Tilde{\lambda}^2, \Tilde{\lambda}^3$ correspond to unit directions for linear forces along the $x$-, $y$-, and $z$-axes, respectively, while $\Tilde{\lambda}^4, \Tilde{\lambda}^5, \Tilde{\lambda}^6$ encode the skew-symmetric representations of moments about these axes.

\section{Structure coefficients of $se(3)$}\label{ap:StructureConstSE3}
The structure coefficients of $se(3)$ capture the underlying (non-)commutative properties of the algebra. These coefficients describe how the differential relationships between basis vectors can be expressed in terms of the algebra’s standard basis.

The interaction between basis elements $e_i$ and $e_j$ can be expressed as a linear combination of basis vectors $e_k$, with the structure coefficients $C^k_{\ ij}$ given by:
\begin{equation}\label{eq:StructureConst}
\frac{\partial e_i}{\partial x^j} - \frac{\partial e_j}{\partial x^i} = C^k_{\ ij} \ e_k.
\end{equation}
Due to the underlying structure of the algebra, the coefficients satisfy the antisymmetry property:
\begin{equation*}
C^k_{\ ij} = - C^k_{\ ji}.
\end{equation*}

We can determine the structure coefficients $C^k_{\ ij}$ using the standard basis of  $se(3)$ (Appendix~\ref{ap:Twists}) together with eq.~\eqref{eq:StructureConst}. The non-zero elements are:
\begin{subequations}
\begin{equation*}
  \begin{split}
  &C^3_{\ 15} = C^1_{\ 26} = C^2_{\ 34} = C^6_{\ 45} = C^4_{\ 56}\\
  &= C^2_{\ 61} = C^3_{\ 42} = C^1_{\ 53} = C^5_{\ 64} = 1;
  \end{split}
\end{equation*}
\begin{equation*}
  \begin{split}
  &C^3_{\ 51} = C^1_{\ 62} = C^2_{\ 43} = C^6_{\ 54} = C^4_{\ 65}\\
  & = C^2_{\ 16} = C^3_{\ 24} = C^1_{\ 35} = C^5_{\ 46}= -1.
  \end{split}
\end{equation*}
\end{subequations}

\section{Left-invariant Kinematic Connection}\label{ap:KinConnection}
To derive the Kinematic Connection, we first express the time rate of change of the body twist $^{\text{B}}\bm{\xi}$ in a form that aligns with the concept of an acceleration screw \citep{stramigioli_geometry_2001}. The acceleration screw describes how the twist evolves over time, incorporating both its direct time derivative and additional terms that arise from the movement of the body-fixed frame:

\begin{equation}\label{eq:AccScrew}
\dfrac{ d (^{\text{B}}\bm{\xi}) }{ dt }=
\begin{pmatrix}
\dot{\bm{v}} \\
\dot{\bm{\omega}}
\end{pmatrix}
+
\begin{pmatrix}
\bm{\omega} \times \bm{v} \\
\bm{0}_{3 \times 1}
\end{pmatrix}.
\end{equation}

The second term on the right-hand side accounts for the fact that the body-fixed frame rotates with angular velocity $\bm{\omega}$, which influences the translational acceleration $\dot{\bm{v}}$. This term is identical to the Coriolis term found in non-inertial reference frames in classical mechanics.

We now express this in terms of the basis components of the twist:
\begin{equation}\label{eq:TimeDerivativeTwist}
\frac{d}{dt} (\xi^k e_k) = \frac{d \xi^k}{dt} e_k + \xi^i \xi^j \ \Gamma^k_{\ ji} \ e_k,
\end{equation}
The first term represents the time derivative of the twist components. The second term introduces correction terms that account for the movement of the body-fixed frame, which are identical to Coriolis and centrifugal terms.

The translational part of the correction term $\xi^i \xi^j \ \Gamma^k_{\ ji} \ e_k$ can be rewritten as $\omega^i \ v^j \ (e_i \times e_j)$. This indexing choice follows from the structure of the Lie algebra  se(3) , where indices $i = \{4,5,6\}$ correspond to rotational basis vectors and indices $j = \{1,2,3\}$  correspond to translational basis vectors. The cross product of a rotational and a translational basis vector results in a translational basis vector.

Since we want to express this in terms of the same basis $e_k$, we rewrite the right-hand side as:
\begin{equation}\label{eq:BasisExpansion}
\omega^i \ v^j \ (e_i \times e_j) = \omega^i \ v^j \ (a^k_{\ ij} \ e_k).
\end{equation}

This leads to an important relationship:
\begin{equation}\label{eq:ExpansionCoefficients}
e_i \times e_j = a^k_{\ ij} \ e_k.
\end{equation}

\subsection{Computing the Expansion Coefficients}
By evaluating eq.~\eqref{eq:ExpansionCoefficients} for the standard basis of $se(3)$ (sec.~\ref{ap:Twists}), we obtain the following relationships for the translational basis:

\begin{subequations}
\begin{equation*}
i = 4, j = 2: e_4 \times e_2 = e_3 \quad \Rightarrow \quad a^3_{\ 42} = 1.
\end{equation*}

\begin{equation*}
i = 4, j = 3: e_4 \times e_3 = -e_2 \quad \Rightarrow \quad a^2_{\ 43} = -1.
\end{equation*}

\begin{equation*}
i = 5, j = 1: e_5 \times e_1 = -e_3 \quad \Rightarrow \quad a^3_{\ 51} = -1.
\end{equation*}

\begin{equation*}
i = 5, j = 3: e_5 \times e_3 = e_1 \quad \Rightarrow \quad a^1_{\ 53} = 1.
\end{equation*}

\begin{equation*}
i = 6, j = 1: e_6 \times e_1 = e_2 \quad \Rightarrow \quad a^2_{\ 61} = 1.
\end{equation*}

\begin{equation*}
i = 6, j = 2: e_6 \times e_2 = -e_1 \quad \Rightarrow \quad a^1_{\ 62} = -1.
\end{equation*}
\end{subequations}

This confirms that the term $(\bm{\omega} \times \bm{v})$ only affects the translational components of eq.~\eqref{eq:AccScrew}, meaning that the correction term exists only for $k = \{ 1, 2, 3 \}$.

\subsection{Derivation of the Christoffel Symbols}
For $i = \{ 4, 5, 6 \}$ and $j = \{ 1, 2, 3 \}$, we equate the right-hand side of eq.\eqref{eq:BasisExpansion} with the last term of eq.\eqref{eq:AccScrew}, yielding:

\begin{equation}\label{eq:CS_a^k_ij}
\Gamma^k_{\ ji} = a^k_{\ ij}.
\end{equation}

For cases where $i = j = \{ 1, 2, 3, 4, 5, 6 \}$, we use the definition of the structure constants:

\begin{equation}\label{eq:StructureCoef_CS}
C^k_{\ ij} = (\Gamma^k_{\ ji} - \Gamma^k_{\ ij}).
\end{equation}

\subsubsection{Translational Christoffel Symbols}
From eq.~\eqref{eq:CS_a^k_ij}, we directly obtain:

\begin{subequations}\label{eq:CSII}
\begin{equation}
\Gamma^3_{\ 24} = \Gamma^1_{\ 35} = \Gamma^2_{\ 16} = 1;
\end{equation}
\begin{equation}
\Gamma^2_{\ 34} = \Gamma^3_{\ 15} = \Gamma^1_{\ 26} = -1.
\end{equation}
\end{subequations}

\subsubsection{Rotational Christoffel Symbols}
The rotational terms are derived from eq.\eqref{eq:StructureCoef_CS} and match the structure constants (appendix~\ref{ap:StructureConstSE3}):

\begin{subequations}\label{eq:CSI}
\begin{equation}
\Gamma^4_{\ 65} - \Gamma^4_{\ 56} = 1 \quad \Rightarrow \quad \Gamma^4_{\ 65} = 0.5, \quad \Gamma^4_{\ 56} = -0.5.
\end{equation}
\begin{equation}
\Gamma^5_{\ 46} - \Gamma^5_{\ 64} = 1 \quad \Rightarrow \quad \Gamma^5_{\ 46} = 0.5, \quad \Gamma^5_{\ 64} = -0.5.
\end{equation}
\begin{equation}
\Gamma^6_{\ 54} - \Gamma^6_{\ 45} = 1 \quad \Rightarrow \quad \Gamma^6_{\ 54} = 0.5, \quad \Gamma^6_{\ 45} = -0.5.
\end{equation}
\end{subequations}

\section{Correction terms}\label{ap:CorrectionTerms}
Applying the Kinematic Connection of appendix~\ref{ap:KinConnection} for body-fixed coordinates, the correction term on the right side of eq.~\eqref{eq:JointSpaceStiffnessCorrection} in matrix form is given by:
\begin{equation}\label{eq:KinConBody}
    \Gamma^k_{\ ij} \ F_k = 
    \begin{pmatrix}
        0 & 0 & 0  & 0 & - F_3 & F_2 \\
        0 & 0 & 0 & F_3 & 0 & -F_1 \\
        0 & 0 & 0 & -F_2 & F_1 & 0 \\
        0 & 0 & 0  & 0 & -\dfrac{F_6}{2} & \dfrac{F_5}{2} \\
        0 & 0 & 0 & \dfrac{F_6}{2} & 0 & -\dfrac{F_4}{2} \\
        0 & 0 & 0 & -\dfrac{F_5}{2} & \dfrac{F_4}{2} & 0
    \end{pmatrix},
\end{equation}
where $(F_1, F_2, F_3)$ are the measured external force components, and $(F_4, F_5, F_6)$ are the external moment components.

If fixed inertial coordinates are chosen to express the Jacobian matrix (denoted with the superscript $F$), a right-invariant connection must be considered instead. This can be achieved by switching the indices $i$ and $j$ of the Christoffel symbols:
\begin{equation}
^{F}\Gamma^k_{\ ji} = \Gamma^k_{\ ij}.
\end{equation}

\section{Example of asymmetric kinematic stiffness}\label{ap:AntiSim_Kkin}

For the Anthropomorphic Arm, as derived in \cite{siciliano_springer_2008}, the rotational part of the Hybrid Jacobian $\bm{J}_r(\bm{q}) \in \mathbb{R}^{3 \times 3}$ 
 \citep{lachner2024explicit} is given by:
\begin{equation}
    \bm{J}_r(\bm{q}) = 
    \begin{pmatrix}
        0 & sin(q_1) & sin(q_1) \\
        0 & -cos(q_1) & -cos(q_1) \\
        1 & 0 & 0
    \end{pmatrix}.
\end{equation}
Its transpose is:
\begin{equation}
    \bm{J}_r(\bm{q})^T = 
    \begin{pmatrix}
        0 & 0 & 1 \\
        sin(q_1) & -cos(q_1) & 0 \\
        sin(q_1) & -cos(q_1) & 0
    \end{pmatrix}.
\end{equation}

Taking the partial derivatives of $\bm{J}_r(\bm{q})^T$ with respect to each joint variable:
\begin{equation}
    \frac{\partial \bm{J}_r(\bm{q})^T}{\partial q_1} = 
    \begin{pmatrix}
        0 & 0 & 0 \\
        cos(q_1) & sin(q_1) & 0 \\
        cos(q_1) & sin(q_1) & 0
    \end{pmatrix},
\end{equation}
\begin{equation}
    \frac{\partial \bm{J}_r(\bm{q})^T}{\partial q_2} = 
    \begin{pmatrix}
        0 & 0 & 0 \\
        0 & 0 & 0 \\
        0 & 0 & 0
    \end{pmatrix},
\end{equation}
\begin{equation}
    \frac{\partial \bm{J}_r(\bm{q})^T}{\partial q_3} = 
    \begin{pmatrix}
        0 & 0 & 0 \\
        0 & 0 & 0 \\
        0 & 0 & 0
    \end{pmatrix}.
\end{equation}
The corresponding kinematic stiffness matrix yields:
\begin{equation}
    \bm{K}_{\text{kin}} =
    \begin{bmatrix}
        \frac{\partial \bm{J}_r(\bm{q})^T}{\partial q_1} \bm{m}_{\text{ext}} &
        \frac{\partial \bm{J}_r(\bm{q})^T}{\partial q_2} \bm{m}_{\text{ext}} &
        \frac{\partial \bm{J}_r(\bm{q})^T}{\partial q_3} \bm{m}_{\text{ext}}
    \end{bmatrix},
\end{equation}
wich finally takes the form:
\begin{equation}\label{eq:KinEx}
    \bm{K}_{\text{kin}} =
    \begin{pmatrix}
        \begin{bmatrix}
            0 & 0 & 0 \\
            cos(q_1) & sin(q_1) & 0 \\
            cos(q_1) & sin(q_1) & 0
        \end{bmatrix} \bm{m}_{\text{ext}} &
        \mathbf{0} &
        \mathbf{0}
    \end{pmatrix}.
\end{equation}
It can be seen that $\frac{\partial \bm{J}_r(\bm{q})}{\partial q_1}$ contains nonzero off-diagonal terms and $\bm{K}_{\text{kin}}$ is asymmetric.

\section{Example of symmetric kinematic stiffness}\label{ap:Sim_Kkin}

The rotational part of the correction term, derived from the Christoffel symbols for the Hybrid Jacobian, is given by (appendix~\ref{ap:CorrectionTerms}):
\begin{equation}
    \bm{\Gamma} \ \bm{m}_{\text{ext}} = 
    \begin{pmatrix}
        0 & \dfrac{m_{\text{ext},3}}{2} & -\dfrac{m_{\text{ext},2}}{2} \\
        -\dfrac{m_{\text{ext},3}}{2} & 0 & \dfrac{m_{\text{ext},1}}{2} \\
        \dfrac{m_{\text{ext},2}}{2} & -\dfrac{m_{\text{ext},1}}{2} & 0
    \end{pmatrix}.
\end{equation}

To demonstrate how this term contributes to the symmetrization of the kinematic stiffness matrix (appendix~\ref{ap:AntiSim_Kkin}), we compute its mapping through the transposed Jacobian:
\begin{equation}\label{eq:MappedGammaEx}
        \bm{J}_r(\bm{q})^T \ (\bm{\Gamma} \ \bm{m}_{\text{ext}}) \ \bm{J}_r(\bm{q})^T =
        \begin{pmatrix}
        0 & A & A \\
        -A & 0 & 0 \\
        -A & 0 & 0
        \end{pmatrix},
\end{equation}
where  A  is defined as:

\begin{equation}
A = \dfrac{1}{2} \left( m_{\text{ext},1} \cos(q_1) + m_{\text{ext},2} \sin(q_1) \right).
\end{equation}

Using this result, we can now verify that the kinematic stiffness matrix, given in eq.~\eqref{eq:KinEx}, becomes symmetric when the correction term is added:
\begin{equation}
    \bm{K}_{\text{kin}} + \bm{J}_r(\bm{q})^T \ (\bm{\Gamma} \ \bm{m}_{\text{ext}}) \ \bm{J}_r(\bm{q})^T = 
        \begin{pmatrix}
        0 & A & A \\
        A & 0 & 0 \\
        A & 0 & 0
        \end{pmatrix}.
\end{equation}

\end{appendices}

\end{document}